\documentclass[10pt,twocolumn,letterpaper]{article}

\usepackage{cvpr}              

\usepackage[acronym]{glossaries}
\usepackage{svg}
\usepackage{soul}
\usepackage{ma
rvosym}

\newacronym{ai}{AI}{Artificial Intelligence}
\newacronym{ann}{ANN}{Artificial Neural Network}
\newacronym{cnn}{CNN}{Convolutional Neural Network}
\newacronym{cots}{COTS}{Commercial-Off-The-Shelf}
\newacronym{dl}{DL}{Deep Learning}
\newacronym{dnn}{DNN}{Deep Neural Network}
\newacronym{dsp}{DSP}{Digital Signal Processing}
\newacronym{fps}{FPS}{Frames Per Second}
\newacronym{flops}{FLOPs}{Floating-point Operations}
\newacronym{hpo}{HPO}{Hyper-Parameter Optimization}
\newacronym{iot}{IoT}{Internet of Things}
\newacronym{mac}{MAC}{Multiply-And-Accumulate}
\newacronym{map}{mAP}{mean Average Precision}
\newacronym{mae}{MAE}{Mean Absolute Error}
\newacronym{mcu}{MCU}{Microcontroller Unit}
\newacronym{ml}{ML}{machine learning}
\newacronym{mpu}{MPU}{microprocessor unit}
\newacronym{mse}{MSE}{mean squared error}
\newacronym{nas}{NAS}{Neural Architecture Search}
\newacronym{nni}{NNI}{Neural Network Intelligence}
\newacronym{qat}{QAT}{Quantization-Aware Training}
\newacronym{soc}{SoC}{System-on-Chip}
\newacronym{sota}{SotA}{state-of-the-art}
\newacronym{ssd}{SSD}{Single Shot Detector}
\newacronym{ee}{EE}{Early Exit}
\newacronym{od}{OD}{Object Detection}
\newacronym{uav}{UAV}{Unmanned Aerial Vehicle}
\newacronym{yolo}{YOLO}{You Only Look Once}

\definecolor{cvprblue}{rgb}{0.21,0.49,0.74}
\usepackage[pagebackref,breaklinks,colorlinks,allcolors=cvprblue]{hyperref}

\def\paperID{*****} 
\def\confName{CVPR}
\def\confYear{2026}

\newcommand{\method}{\textsc{BlankSkip}\xspace}

\title{\method: Early-exit Object Detection onboard Nano-drones}

\author{
  Carlo Marra$^\dagger$$^*$, 
  Beatrice Alessandra Motetti$^\dagger$, 
  Alessio Burrello$^\dagger$, 
  Enrico Macii$^\dagger$,\\
  Massimo Poncino$^\dagger$, 
  Daniele Jahier Pagliari$^\dagger$ \\[0.3em]
  {$^\dagger$Politecnico di Torino, Turin, Italy}\\
  \normalsize{$^*$Corresponding author, email: \texttt{carlo.marra@polito.it}}
}

\begin{document}

\maketitle

\begin{abstract}
Deploying tiny computer vision Deep Neural Networks (DNNs) on-board nano-sized drones is key for achieving autonomy, but is complicated by the extremely tight constrained of their computational platforms ($\approx 10\,\mathrm{MiB}$ memory, $\approx 1\,\mathrm{W}$ power budget). Early-exit adaptive DNNs that dial down the computational effort for ``easy-to-process'' input frames represent a promising way to reduce the average inference latency. However, while this approach is extensively studied for classification, its application to dense tasks like object detection (OD) is not straight-forward. In this paper, we propose \method, an adaptive network for on-device OD that leverages a simple auxiliary classification task for early-exit, i.e., identifying frames with no objects of interest.
With experiments using a real-world nano-drone platform, the Bitcraze Crazyflie 2.1, we achieve up to 24\% average throughput improvement with a limited 0.015 \gls{map} drop compared to a static MobileNet-SSD detector, on a state-of-the-art nano-drones OD dataset. 

\end{abstract}

\section{Introduction}

Nano-drones are small \glspl{uav} that enable unique applications such as indoor exploration, infrastructure inspection, and search-and-rescue operations, where their agility and small form factor (diameter smaller than 10\,cm) open scenarios inaccessible to larger systems~\cite{lifetime_nano_blimps}. Operating autonomously in these settings requires onboard perception: for instance the drone must be able to detect and localize objects in real time, without relying on remote computation offloading, as the latency and communication overhead introduced would be incompatible with fast, reactive flight. However, the hardware aboard nano-drones is severely constrained by strict power and weight budgets. Platforms such as the Bitcraze Crazyflie 2.1~\cite{cf21_datasheet}, equipped with the GAP8 \gls{soc} \gls{ai} deck~\cite{gap8}, offer only a few hundred MHz of clock frequency, less than 10 MiB of total on-board RAM, and operate under a total power envelope of a few hundred milliwatts for the compute part~\cite{himax}. Under these constraints, even compact \gls{dnn} architectures designed for embedded devices deliver low frame rates, that leave little margin for the multitasking required by autonomous flight (e.g., simultaneous navigation, obstacle avoidance, and target detection). Existing approaches to onboard \gls{od}, in particular, have focused on compressing \glspl{dnn} through quantization, pruning, and lightweight architectural design~\cite{howard2017mobilenets, sandler2018mobilenetv2, wong2019yolonano}, yielding models that fit within the memory and compute budget of embedded processors. Yet these solutions share a fundamental limitation: they apply a fixed computational graph to every input frame, regardless of its content. This is particularly wasteful in nano-drone exploration scenarios, where a large fraction of frames may contain little to no information. Adaptive inference methods~\cite{yu2019slimmable, teerapittayanon2016branchynet} address this inefficiency by adjusting compute at runtime in an input-dependent way. Among them, \gls{ee} \glspl{dnn} have shown promising results in reducing average latency for image classification~\cite{ee-survey}. However, their application to \gls{od}, and to severely constrained platforms such as nano-drones, remains largely unexplored.

In this work, we propose \method, an \gls{ee} mechanism for \gls{od} on nano-drones that exploits the sparsity of object presence across frames. Rather than applying the full detection pipeline to every input, we decompose the problem into two sequential sub-tasks: a lightweight binary classifier, attached at an intermediate layer of the backbone, first determines whether a frame contains any object of interest; only if so, inference continues through the full detection head. This yields a reduction in average latency proportional to the fraction of empty frames in the scene, directly translating into higher effective frame rates and freed computational budget for concurrent tasks.

More in detail, the following are our main contributions:
\begin{itemize}
    \item We train \method end-to-end with a composite loss that jointly optimizes detection quality and early-exit reliability, with class weights enabling explicit control over the missed-detection vs. computational savings trade-off.
    \item We formulate the joint optimization of branch placement, training hyperparameters, and confidence threshold as a Bayesian \gls{hpo} problem.
    \item We evaluate \method on two datasets against a static (non-adaptive) baseline: on Himax EE, a nano-drone-specific dataset from~\cite{himax}, we achieve a \gls{map} of 0.593, matching the result of a static baseline using the same backbone from the original paper (0.591) while skipping 39.8\% of frames and reducing the average \gls{mac} operations per frame by 22.5\%. On the more challenging Cityscapes dataset \cite{cityscapes}, we maintain competitive \gls{map} (0.204 vs. 0.229, -11\%) while skipping 26.7\% of frames and reducing the average \glspl{mac} by 12.1\%.
    \item We deploy an 8-bit quantized version of our model on the Bitcraze Crazyflie 2.1 GAP8 SoC, achieving a \gls{map} of 0.588 at 1.86 average \gls{fps}, a 24\% throughput improvement over the baseline (1.50 FPS), with 0.015 \gls{map} drop.
\end{itemize}

\section{Background and Related Works}
\label{sec:background}

\subsection{Nano-drone computational platforms}

As our target for deployment, we consider the Bitcraze Crazyflie 2.1, a \gls{cots} quadrotor with a diameter of 10\,cm and a weight of 27\,g~\cite{cf21_datasheet}. The drone is operated by an STM32F405 \gls{mcu} dedicated to state estimation and actuation control. The platform can be extended with companion boards, most notably the 
AI-deck, which embeds a GAP8 \gls{soc} from GreenWaves Technologies~\cite{gap8}. GAP8 is a parallel ultra-low-power processor based on the RISC-V architecture, designed for \gls{dsp} and \gls{ai} applications, featuring one fabric controller core and a cluster of 8 additional cores clocked up to 175\,MHz, designed for parallel workloads acceleration. The on-chip memory hierarchy consists of 64\,kB of L1 scratchpad shared among the cluster cores and 512\,kB of L2 SRAM on the fabric controller side. 
The AI-deck further provides a Himax HM01B0 ultra-low-power grayscale camera and off-chip L3 memories (8\,MB RAM and 64\,MB Flash) accessible via HyperBus.

\subsection{Object detection on constrained devices} 

Deploying \gls{od} on resource-constrained devices has attracted growing research interest, driven by the demand for real-time onboard perception in mobile and embedded systems. The \gls{ssd}~\cite{liu2016ssd} has become a widely adopted framework for this setting, consisting of a feature extraction backbone and multiple convolutional prediction heads. MobileNets~\cite{howard2017mobilenets,sandler2018mobilenetv2} are commonly used as low-cost backbones for \gls{ssd} in embedded contexts. Compact end-to-end detectors such as \gls{yolo} Nano~\cite{wong2019yolonano} have also been explored, leveraging \gls{nas} driven design to minimize both model size and inference cost; while \gls{ssd}-based models offer a more compelling speed advantage, \gls{yolo}-based alternatives usually trade some throughput for improved detection accuracy and robustness \cite{yoloVSssd}.

These architectures are typically further compressed for deployment via post-training quantization~\cite{jacob2018quantization} and structured pruning~\cite{han2015learning}. Nevertheless, deploying such models on severely constrained platforms remains challenging: the work of ~\cite{tranquangkhoi2021} reports only 0.71 FPS running SSDLite-MobileNetV2 on a Raspberry Pi B3+ mounted on a standard-sized drone, already exceeding the power and size budget of nano-drone systems. 
The authors of~\cite{lamberti2021lowpower} demonstrate a more power-efficient deployment on a GAP8 \gls{mcu}, achieving 1.6 FPS at 117\,mW, yet relying on a fixed computational graph regardless of input content.

\textbf{Adaptive inference.} The goal of adaptive inference is to dynamically adjust the computational cost at runtime based on the complexity of each input \cite{dynamic-net-survey}. Several directions have been explored: Slimmable Networks~\cite{yu2019slimmable} allow a single model to operate at different channel widths, while cascade approaches route easy inputs through lightweight models and hard inputs through heavier ones~\cite{dynamicdet2023}. In \gls{ee} models, side-branch classifiers are inserted at intermediate layers, and inference is terminated as soon as a confidence criterion, typically entropy-based, is met. 

BranchyNet~\cite{teerapittayanon2016branchynet} introduced the \gls{ee} paradigm, demonstrating that a large fraction of inputs can be correctly classified at shallow layers with significant latency savings. A key advantage of \gls{ee}, which makes this solution deployment-friendly, is that input-adaptive compute is achieved within a single model, without the memory overhead of multiple networks, except for the extra branch, which is usually negligible. While this is also true for Slimmable networks ~\cite{yu2019slimmable}, the latter introduce additional complexity during inference (extra routing logic, partial loading of each layer's parameters, which influences optimal computation scheduling, etc).

However, while \gls{ee} has been extensively studied for image classification~\cite{teerapittayanon2016branchynet, yu2019slimmable, ee-survey}, its application to dense tasks like \gls{od} remains largely underexplored. This is mainly because entropy-based \gls{ee} criteria designed for classification do not naturally extend to tasks such as \gls{od}, which produce dense outputs (grids of bounding box locations and class probabilities), each with its own score distribution.

AdaDet~\cite{adadet2023} addresses this gap by integrating detection heads at intermediate backbone layers and introducing a bounding-box uncertainty criterion to assess exit confidence, while DynamicDet~\cite{dynamicdet2023} adopts a cascade of two detectors with a learned difficulty router. These works, however, target high-power GPU platforms and are incompatible with the tight computational constraints of nano-drone deployment. While they do not report exact model sizes, their simpler models require $\approx$5-10\,GFLOPs per inference, roughly one order of magnitude more than the \glspl{dnn} we deploy onboard the Crazyflie in this paper.

Sabet et al.~\cite{sabet2022temporal} propose temporal early-exits for video object detection, where changes between consecutive frames trigger full computation while unchanged frames reuse previous detections.

This approach relies on fixed camera placement (the authors target surveillance scenarios), therefore, it does not apply to drone use cases.
Hector et al.~\cite{hector2021scalable} introduce elastic neural networks for edge deployment, incorporating multiple detection heads at different depths to enable runtime adaptation based on external resource constraints (e.g., battery levels). However, this method relies on an exogenous signal rather than per-sample content-based decisions to tune the computational effort, which limits applicability to scenarios with highly variable frame content such as nano-drone exploration.
\section{Methodology}
\label{sec:methodology}

In nano-drone applications, \gls{od} networks must process video streams in real-time under severe power, memory, and compute constraints. However, a significant fraction of frames captured during exploration may contain no objects of interest. Therefore, processing these ``empty'' frames unnecessarily wastes compute and, therefore, latency. Based on this intuition, we propose \method, a simple yet effective \gls{ee} mechanism that classifies frames as empty before completing the full detection pipeline. 

\method reduces the average \gls{od} latency, enabling higher average frame rates. This, in turn, can improve detection rates during fast flight maneuvers and, consequently, enable more efficient exploration of the environment. For example, \cite{himax} has analyzed the non-trivial dependencies between flight speed, inference latency and overall exploration time for nano-drones, showing that faster-moving \glspl{uav} often ``miss'' objects due to the low \gls{od} rates. Additionally, in multi-task deployments, freed computational resources thanks to \gls{ee} can be allocated to other, non-critical, concurrent tasks.

\subsection{Problem formulation}

We reformulate \gls{od} as two sequential sub-problems:
\begin{enumerate}
    \item \textit{Binary Classification}: Determine whether an image contains objects of interest (non-empty) or not (empty);
    \item \textit{Object Detection}: Execute full detection only when the image is confidently classified as non-empty.
\end{enumerate}

A straightforward approach would involve two separate models, one per stage. However, this would greatly increase the memory footprint and total computational cost of the system. Instead, we share the same backbone network for both tasks. To achieve this, we adapt the \gls{ee} paradigm, originally proposed in \cite{teerapittayanon2016branchynet}: an intermediate branch performs binary empty/non-empty classification to gate the \gls{od} pipeline. When an image is confidently classified as empty, the system bypasses the remaining backbone layers and detection stages entirely, yielding the computational savings described above while maintaining a single unified model.

Let $I \in \mathbb{R}^{H \times W \times C}$ denote an input image and $\mathcal{D}(\cdot)$ a conventional \gls{od} pipeline composed of a feature extraction backbone and detection heads. At intermediate layer $\ell$ of the backbone, we extract feature maps $\mathbf{x}_\ell \in \mathbb{R}^{H' \times W' \times C'}$ and pass them to a lightweight \gls{ee} branch $f_{\text{EE}}(\cdot)$ that produces binary classification probabilities:
\begin{equation}
P(c \mid \mathbf{x}_\ell) = \text{softmax}(f_{\text{EE}}(\mathbf{x}_\ell)), \quad c \in \{0, 1\}
\end{equation}
where $c=1$ denotes ``empty'' and $c=0$ denotes ``non-empty''. The inference decision mechanism is:
\begin{equation}\label{eq:ee_mechanism}
\mathcal{O}(I) =
\begin{cases}
\emptyset & \text{if } P(1 \mid \mathbf{x}_\ell) \geq \tau \\
\mathcal{D}(I) & \text{otherwise}
\end{cases}
\end{equation}
where $\tau \in [0.5,1]$ is a confidence threshold controlling the trade-off between computational savings and detection reliability. High-confidence gating (high $\tau$) minimizes false positives, i.e. incorrectly skipping images containing objects, which is critical as it would result in missed detections. Conversely, false negatives (empty images classified as non-empty), incur a computational overhead but do not necessarily affect detection accuracy, as $\mathcal{D}(\cdot)$ can still produce empty outputs when no objects are present.

\subsection{Network architecture}

\begin{figure*}[t]
    \centering
    \includegraphics[width=0.9\textwidth]{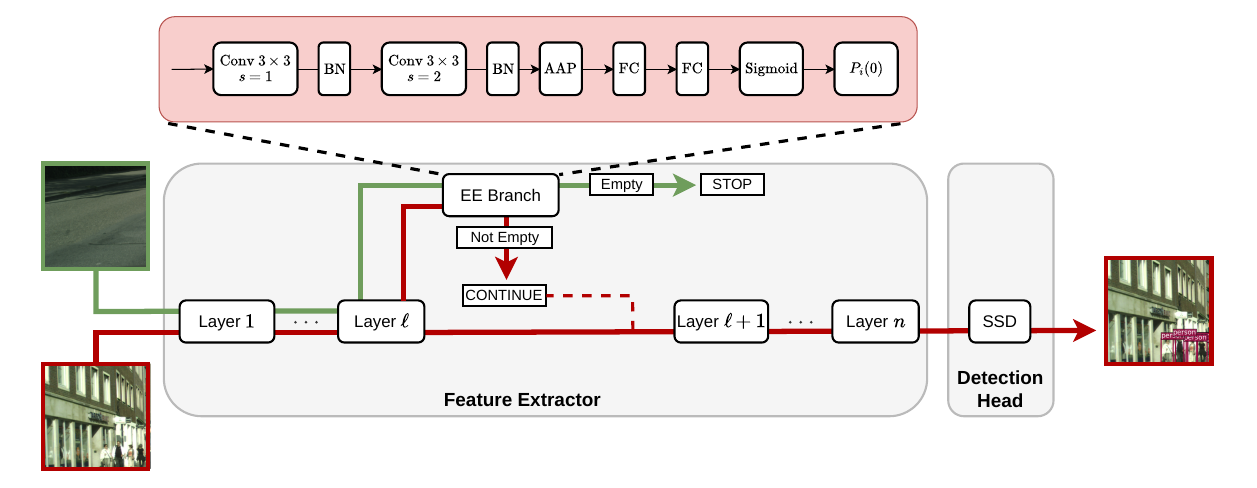}
    \caption{Overview of the proposed \method architecture. The backbone processes the input frame through layers $1$ to $\ell$. Then, the \gls{ee} branch predicts the ``emptyness" probability $P_i(1)$. If the scene is classified as \textit{empty}, inference stops early (green path), avoiding the remaining backbone layers and the \gls{ssd} detection head. Otherwise, the full pipeline runs to produce bounding box predictions (red path).}
    \label{fig:ee_net_schema}
\end{figure*}

Our \gls{od} pipeline is based on a \gls{ssd} algorithm composed of a MobileNetV2 feature extractor and multiple detection heads \cite{sandler2018mobilenetv2, liu2016ssd}. As illustrated in Fig.~\ref{fig:ee_net_schema}, the \gls{ee} branch attaches at an intermediate layer $\ell$ of the backbone, which shall balance sufficient semantic abstraction for reliable empty/non-empty discrimination with the computational savings obtained when early-exit triggers.

The \gls{ee} branch $f_{\text{EE}}(\cdot)$ is deliberately lightweight to minimize overhead. Its architecture consists of:

\begin{enumerate}
    \item Two $3 \times 3$ convolutional layers with batch normalization and ReLU6 activation, where the second uses stride 2;
    \item Global Average Pooling (GAP) to collapse spatial dimensions into a compact feature vector;
    \item Two fully-connected layers.
\end{enumerate}

Formally, the \gls{ee} branch computes:
\begin{equation}
\mathbf{h} = \text{GAP}(\sigma(\text{BN}(W_2 * \sigma(\text{BN}(W_1 * \mathbf{x}_\ell)))))
\end{equation}
\begin{equation}
f_{\text{EE}}(\mathbf{x}_\ell) = W_4(\sigma(W_3 \mathbf{h}))
\end{equation}
where $*$ denotes convolution and $\sigma$ is ReLU6. The branch dimensions are designed such that the first convolutional layer maintains the channel dimension of $\mathbf{x}_\ell$, while the second reduces it to 64 channels. The two fully-connected layers produce 64-dimensional intermediate feature vectors, before the final 2-class output. The \gls{ee} branch introduces 50K--138K additional parameters depending on the channel dimension at layer $\ell$, increasing the static MobileNetV2 SSD parameter count by only 1.1--3.0\% over the original 4.67M. Its computational cost in \gls{mac} operations varies with branch placement and the spatial resolution of the feature map, as detailed below.

\subsection{Training strategy}

The system is trained end-to-end with a composite loss:
\begin{equation}\label{eq:composite_loss}
\mathcal{L}_{\text{total}} = \mathcal{L}_{\text{loc}} + \mathcal{L}_{\text{cls}} + \lambda \cdot \mathcal{L}_{\text{EE}}
\end{equation}
where $\mathcal{L}_{\text{loc}}$ is Smooth-L1 localization loss, $\mathcal{L}_{\text{cls}}$ is classification loss \cite{liu2016ssd}, and $\mathcal{L}_{\text{EE}}$ is an additional weighted binary cross-entropy term penalizing empty/non-empty misclassifications by the \gls{ee} branch. Formally:

\begin{equation}
\mathcal{L}_{\text{EE}} = -\frac{1}{N}\sum_{i=1}^{N} \left[w_0 \cdot (1-y_i) \log P_i(0) + w_1 \cdot y_i \log P_i(1)\right]
\end{equation}
where $P_i(c) = P(c \mid \mathbf{x}_\ell^{(i)})$ and $y_i \in \{0,1\}$ indicates whether image $i$ is empty. The class weights $w_0$ and $w_1$ enable tuning the trade-off between detection reliability and computational efficiency. Higher $w_0$ penalizes false positives (non-empty images incorrectly classified as empty), prioritizing detection completeness, which is critical when missing objects are costly. Conversely, a higher $w_1$ penalizes false negatives (empty images unnecessarily processed), maximizing energy savings in resource-constrained scenarios. The weighting parameter $\lambda$ in Eq.~\ref{eq:composite_loss} balances the \gls{ee} objective with detection accuracy. Ground truth labels are derived automatically from detection annotations, requiring no additional annotation effort.

\subsection{Architectural exploration}
\label{subsec:bayopt}

The design of an \gls{ee} network involves several interdependent decisions that cannot be optimized in isolation. The placement of the \gls{ee} at layer $\ell$ affects both the semantic richness of features available for classification and the potential computational savings  \cite{teerapittayanon2016branchynet, adadet2023}. 
The relationship between branch placement and compute cost is non-trivial: placing it earlier enables larger potential backbone savings, but the branch must process high-resolution feature maps, thus being more expensive in terms of \gls{mac}s; conversely, deeper layers have smaller spatial dimensions, making the branch usually lighter in terms of computation, but leave less backbone computation to skip. Additionally, later layers encode semantically richer representations, potentially improving the branch's ability to discriminate between empty and non-empty frames. This creates the potential for a sweet spot within the network where branch cost, compute savings, and classification accuracy are jointly optimized.
Moreover, the optimal branch placement influences training dynamics, and thus the optimal learning rate, loss weights, etc.

This interdependency extends to the confidence threshold $\tau$, another critical design parameter. While the threshold can be adjusted at inference time to dynamically tune the accuracy/compute trade-off, its optimal value is coupled with training hyperparameters.
We address this multi-dimensional problem through a Bayesian \gls{hpo} stage \cite{bayesian}. We select a Bayesian Optimization method because our search space contains mixed parameter types, including categorical branch placement, continuous learning rates, and loss weights. Moreover, each objective evaluation requires full network training, making exhaustive search computationally prohibitive.

We employ the Tree-structured Parzen Estimator (TPE) \cite{tpe} implemented in Optuna \cite{optuna} to search over candidate EE branch placements, learning rates, batch sizes, loss weights, and class weights. The composite \gls{hpo} objective function balances three competing goals: preserving detection quality (measured by mAP without early-exit to isolate detection performance), maximizing computational savings, and achieving high EE classification accuracy. Formally, we maximize:

\begin{equation}\label{eq:hpo}
\mathcal{J} = \text{mAP}_{\text{baseline}} \times \mathcal{S}(\ell) \times A_{\text{EE}}(\ell)
\end{equation}
where $\text{mAP}_{\text{baseline}}$ is the detection performance without early-exit. $\mathcal{S}(\ell) = 1 - \mathcal{F}_{\text{EE}}(\ell)/\mathcal{F}_{\text{full}}$ represents the static \gls{flops} saving achievable by exiting at layer $\ell$, with $\mathcal{F}_{\text{EE}}(\ell)$ and $\mathcal{F}_{\text{full}}$ denoting the computational cost of early-exit and full forward pass, respectively. Lastly, $A_{\text{EE}}(\ell)$ is the EE branch classification accuracy at the optimal threshold. For each candidate configuration, we perform full network training followed by post-training \gls{ee} threshold ($\tau$) optimization via ternary search.

\subsection{Deployment details}

To deploy baseline (static) MobileNet-SSD detectors and \method \gls{ee} solutions on the Floating Point Unit-less GAP8 \gls{soc}, we quantize all \glspl{dnn} to 8-bit integer format with \gls{qat}, using the open-source PLiNIO library~\cite{plinio}. We apply affine quantization, with simple min-max quantizers for weights and PaCT~\cite{choi2018pact} for activations. We deploy full-integer models, folding normalization layers parameters in the preceding linear/convolutional layers. 
We use the open-source DORY compiler~\cite{dory} to generate low-level C code for GAP8, accelerating all supported layers on the 8-core parallel cluster.
\section{Experimental Results}
\label{sec:results}

\subsection{Experimental setup \& datasets}\label{sec:setup}
We compare \method with the work of Lamberti et al.~\cite{himax}, which trained static MobileNet-SSDs to perform object detection while navigating an environment through different bio-inspired algorithms, targeting the same nano-\gls{uav} platform considered in our work. In addition, to show that our method generalizes to larger and more complex datasets, we also evaluate it on a version of Cityscapes~\cite{cityscapes} adapted for \gls{od} and for our device class.

\textbf{Himax.}
The Himax dataset~\cite{himax} contains 321 training and 279 test images at $240 \times 320$ resolution, collected with the on-board camera of the Crazyflie 2.1 \gls{uav}, and annotated for two object categories: "Bottle" and "Tin can". Since all original images contain at least one object, the dataset does not reflect the high proportion of empty frames typical of real-world deployment scenarios. To address this, we augment the dataset with synthetically generated negative samples. Specifically, for each image we extract random object-free regions ($40$--$70\%$ of the image area), verified by zero intersection with all bounding boxes; positive examples are generated analogously by generating a randomly centered crop around one of the objects in the original dataset. All crops are resized to $240 \times 320$ via letterbox scaling. The augmented training set comprises 496 images (321 positive, 175 negative), and the test set 462 images (279 positive, 183 negative). We refer to this augmented dataset as Himax \gls{ee}, to distinguish it from the original Himax dataset, on which we also report results for direct comparison with \cite{himax}.

\textbf{Cityscapes.}
Cityscapes is an urban scene understanding dataset originally designed for semantic and instance segmentation, covering eight object categories: "Person", "Rider", "Car", "Truck", "Bus", "Train", "Motorcycle", and "Bicycle". To adapt it for \gls{od}, we derive bounding box annotations directly from the instance masks. The original images are at $2048 \times 1024$ resolution, which makes one-shot processing unfeasible on nano-\gls{uav} class devices. In fact, the peak intermediate activations of a static MobileNet-SSD would reach 62.88 MiB, which is almost 1 order of magnitude more than the total RAM available in GAP8. A common solution to enable inference with such high-resolution images on memory-limited targets is sequential patch-based processing~\cite{patch-based-inf}, at the cost of increased latency and of a potential \gls{map} drop for objects crossing the patch boundaries.

We follow this approach, partitioning each image into eight non-overlapping $512 \times 512$ tiles, thus reducing the peak memory to 7.86 MiB, which fits the GAP8 available RAM. The resulting dataset comprises 27,800 tiles (19,040 train\,/\,4,760 validation\,/\,4,000 test). Notably, approximately $36\%$, $32\%$, and $24\%$ of the train, validation, and test tiles are empty, not requiring extra negative sampling for \gls{ee} training. This suggests that \gls{ee} can be beneficial not only when processed frames are frequently empty, but also in scenarios with denser object distributions, if images are processed in patches.

\textbf{Implementation details.}
All models are implemented in PyTorch 2.10. We generate all \method models attaching \gls{ee} branches to a MobileNet-SSD network with a width multiplier $\alpha =$ 1.0$\times$. For Himax and Himax EE experiments, the weights of the backbone are initialized from a detector pre-trained on an Open Images~\cite{OpenImages} subset covering the Bottle and Tin can categories, as proposed by the dataset authors~\cite{himax}. For Cityscapes experiments, weights are initialized from a COCO~\cite{coco} pre-trained detector.

\subsection{Himax EE results}

\begin{table*}[t]
\centering
\caption{Effect of \gls{ee} branch positioning on Himax EE (float32).}
\label{tab:himax_ee_details}
\small
\begin{tabular}{lccccccc}
\toprule
Model & $\tau$ & mAP & mAP$_{\text{no-EE}}$ & Acc [\%] & Skip [\%] & \gls{mac}$_{\text{avg}}$ [M] & \gls{mac}$_{\text{EE}}$ [M] \\
\midrule
Static 0.5$\times$  & -- & 0.412 & -- & -- & -- & 193 & -- \\
Static 0.75$\times$ & -- & 0.479 & -- & -- & -- & 358 & -- \\
Static 1.0$\times$  & -- & 0.591 & -- & -- & -- & 534 & -- \\
\midrule
\gls{ee}, Stage 1 ($\ell=1$)          & 0.541  & 0.543 & 0.603 & 85.3 & 34.2 & 708 & 375 \\
\gls{ee}, Stage 2 ($\ell=2$)          & 0.717  & 0.533 & 0.587 & 89.6 & 39.2 & 456 & 178 \\
\gls{ee}, Stage 3 ($\ell=4$)          & 0.742  & 0.555 & 0.596 & 89.6 & 38.7 & 406 & 165 \\
\textbf{\gls{ee}, Stage 4 ($\ell=9$)} & \textbf{0.685} & \textbf{0.593} & \textbf{0.616} & \textbf{94.4} & \textbf{39.8} & \textbf{414} & \textbf{230} \\
\gls{ee}, Stage 5 ($\ell=11$)         & 0.904  & 0.592 & 0.620 & 95.7 & 41.1 & 430 & 266 \\
\bottomrule
\end{tabular}
\\[2pt]
{\footnotesize \gls{mac}$_{\text{avg}}$: dataset average. \gls{mac}$_{\text{EE}}$: early-exit path cost. mAP$_{\text{no-EE}}$: mAP with early-exit disabled.}
\end{table*}

We compare our EE solution with three static MobileNetV2-SSD configurations with width multipliers $\alpha \in \{0.5, 0.75, 1.0\}$. The full-width model (SSD-MbV2-1.0) has 4.67 M parameters and requires 534 M\gls{mac}s per forward pass, while SSD-MbV2-0.75 and SSD-MbV2-0.5 have 2.68 M and 1.34 M parameters and require 358 M\gls{mac}s and 193 M\gls{mac}s, respectively.

All baseline models are trained following the 
recommended hyperparameters configuration from~\cite{himax}: 100 epochs, RMSprop optimizer with an initial learning rate of $1 \times 10^{-4}$ and a batch size of 24. The learning rate is decayed by a factor of 0.95 every 25 epochs.  Data augmentation includes random horizontal mirroring, random crop sampling, and random brightness adjustment, individually applied with a probability of 0.5.
We first verify our training setup on the original Himax dataset (without empty frames), reproducing their baseline results: our static MobileNetV2-SSDs achieve a \gls{map} of 0.571, 0.495, and 0.429 at width multipliers $1.0\times$, $0.75\times$, and $0.50\times$, respectively, similar or better than the 0.550, 0.460, and 0.430 reported in the original work.
On Himax~\gls{ee}, which includes empty frames, the static MobileNetV2-SSDs achieve a \gls{map} of 0.591, 0.479 and 0.412 respectively, showing that the augmented dataset exhibits similar complexity to the original one.

For \method models, in order to analyze in detail the impact of \gls{ee} branch placement on computational cost and accuracy, we perform 5 separate Bayesian \gls{hpo} runs, one for each of the first 5 stages of the MobileNetV2 1.0$\times$ backbone. Each stage comprises one or more inverted residual block layers, sharing the same output resolution and channel count. Namely, stages $(1,2,3,4,5)$ include $(1,2,3,4,3)$ layers respectively. Each run forces the \gls{ee} branch to be attached to one of the layers belonging to that stage, and optimizes five hyperparameters: the specific \gls{ee} layer within the stage, the branch learning rate, the batch size, the loss weight $\lambda$ (Eq.~\ref{eq:composite_loss}), and the class weight $w_1$ for non-empty samples. Each optimization run performs 100 trials, maximizing the composite objective presented in Eq.~\ref{eq:hpo}. For each candidate configuration, the confidence threshold $\tau$ is selected via ternary search on validation data to maximize overall classification accuracy.

Table~\ref{tab:himax_ee_details} reports detailed metrics for the five \gls{ee} configurations discovered through Bayesian optimization, in float32 precision. The table reports mAP and mAP$_{\text{no-EE}}$, representing detection accuracy with and without early-exit enabled, respectively. Acc denotes the \gls{ee} branch classification accuracy, while Skip indicates the percentage of images triggering early-exit. \gls{mac}$_{\text{avg}}$ shows the average computational cost across the dataset with \gls{ee} active, and \gls{mac}$_{\text{EE}}$ represents the cost of the early-exit path when taken.

Notably, all \gls{ee} configurations achieve a mAP$_{\text{no-EE}}$ higher than the static full-width model (0.591). We attribute this improvement to the fact that the \gls{ee} branch likely provides beneficial gradient signals to intermediate layers during training, acting as a regularizer similarly to what observed in BranchyNet~\cite{teerapittayanon2016branchynet}.

Stage~4 ($\ell=9$) emerges as the most accurate configuration, achieving a mAP of 0.593 with an average cost of 414~M\gls{mac}s per image. The 94.4\% \gls{ee} branch accuracy enables skipping full detection for 39.8\% of the images. Compared to a static full-width baseline (534~M\gls{mac}s), Stage~4 reduces average inference cost by 22.5\% while maintaining competitive detection accuracy. Crucially, when early-exit is taken, the cost drops to only 230~M\gls{mac}s, a 57.9\% reduction compared to the baseline.

Interestingly, the results show that placing the \gls{ee} branch in shallow layers (stage 1 or 2) is significantly less accurate (85.3-89.6\%). This demonstrates that identifying frames with no objects of interest, without relying on priors on the stillness of the frame (as in~\cite{sabet2022temporal}) is a non-trivial task that requires latent representations with sufficient semantic abstraction, thus justifying an \gls{ee} approach versus other forms of skipping. Additionally, Stage 1 placement also suffers from prohibitively high early-exit costs due to processing large feature maps at shallow depths. Despite a 34.2\% skip rate, the average cost exceeds the static baseline (708 M\gls{mac}s), resulting in a net computational overhead rather than savings. Moreover, the low \gls{ee} classification accuracy at earlier stages also further degrades the overall \gls{map}.

The limited skip rates (35--41\%) across all configurations reflect the dataset composition: with approximately 40\% negative samples in the test set, most images require full detector processing. Nevertheless, the results validate \method's effectiveness in preserving detection quality while reducing inference cost, especially considering that, in real-world settings, the share of empty images could be substantially higher.

\begin{figure}[t]
\centering
\includegraphics[width=\columnwidth]{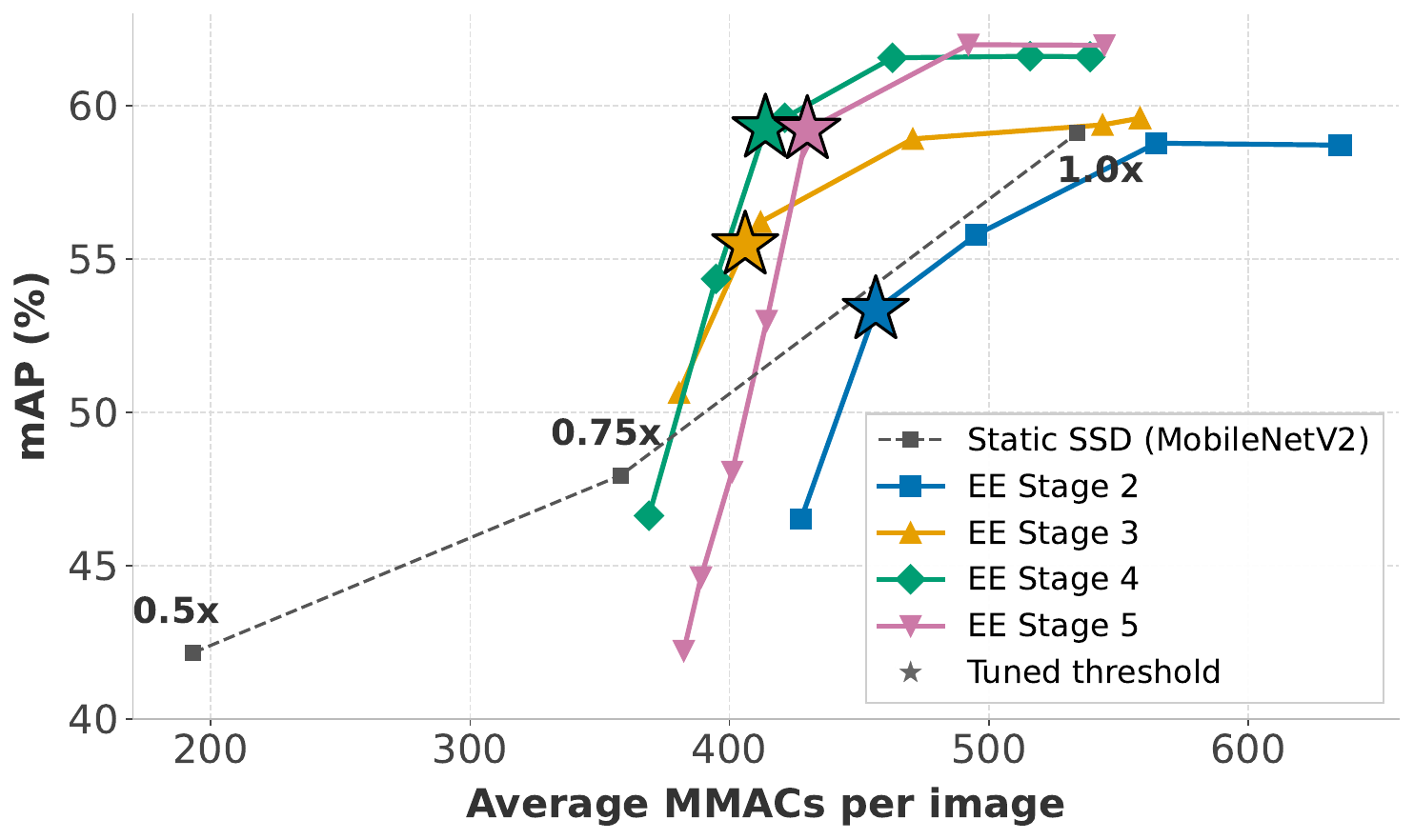}
\caption{Accuracy-efficiency trade-off for with threshold $\tau$ sweep (0.5-0.99) on the Himax EE dataset. Stars mark optimal thresholds according to Eq.~\ref{eq:hpo}. Dashed line: static SSD at different widths. Stage 1 is omitted due to prohibitively high exit costs from large feature maps (120$\times$160).}
\label{fig:himax_pareto}
\end{figure}

Figure~\ref{fig:himax_pareto} visualizes the accuracy-efficiency Pareto frontier obtained across stages 2-5 by sweeping the \gls{ee} decision threshold $\tau$ of Eq.~\ref{eq:ee_mechanism} in the [0.5,0.99] range, with stars marking the optimal operating points selected by Bayesian Optimization to maximize Eq.~\ref{eq:hpo} and reported in Table~\ref{tab:himax_ee_details}. Points to the right of each curve correspond to \textit{larger} $\tau$ values, that let the system skip detection only if the \gls{ee} confidence is very high, thus yielding higher-cost, higher-mAP results. Conversely, points to the left are obtained with \textit{small} $\tau$, thus achieving lower-mAP and lower-cost.
The dashed line represents the static SSD Pareto frontier at different width multipliers. For better visualization, we omit stage 1 entirely due to the sub-optimal results discussed above, and we do not show points reaching a mAP smaller than the static 0.5$\times$ architecture.

The threshold sweep reveals further insights into the behavior of each configuration. The \gls{map} obtained placing the branch at Stage~5, despite being competitive at the optimum according to Eq.~\ref{eq:hpo}, degrades more rapidly as $\tau$ decreases. Stage~2, while offering the lowest early-exit path cost (178~M\gls{mac}s), is unable to match the accuracy of deeper configurations even at high thresholds, confirming that shallow representations are insufficient for reliable empty-frame detection.
Based on this analysis, we select Stage 4 for \gls{qat} and deployment.

Note that, once the \gls{ee} branch placement is selected, different operating modes (e.g., different points from the green curve in Fig.~\ref{fig:himax_pareto}) can still be switched while the system is operating, by changing a single scalar configuration ($\tau$), thus enabling runtime tuning of the latency vs mAP trade-off.
 
\subsubsection{Deployment results}

\begin{table*}[t]
\centering
\caption{Comparison of static MobileNetV2-SSD configurations and \method on the Himax EE dataset.
         Int8 quantization is applied to all models. Deployment metrics are measured on GAP8.}
\label{tab:himax_qat}
\small
\begin{tabular}{lcccccc}
\toprule
Model & mAP & Operations & Efficiency & Latency & Throughput & Model Size \\
      &     & (M\gls{mac}s)    & (\gls{mac}/cycle) & (ms)   & (FPS)      & (MB) \\
\midrule
Static MbV2-SSD 0.5$\times$  & 0.415 & 193  & 4.14   & 285.3    & 3.50   & 1.16 \\
Static MbV2-SSD 0.75$\times$ & 0.498 & 358  & 4.22 & 523.6 & 1.91 & 2.53 \\
Static MbV2-SSD 1.0$\times$  & 0.555 & 534  & 4.96 & 666.7 & 1.50 & 4.41 \\
\midrule
\gls{ee}, Stage 4 ($\ell=9$)           & 0.540 & 414  & 4.87 & 537.1 & 1.86 & 4.49 \\
\bottomrule
\end{tabular}
\end{table*}

To validate the practical feasibility of our approach on nano-drone hardware, we deploy the optimized Stage 4 \method model alongside static baseline configurations (1.0$\times$, 0.75$\times$, and 0.5$\times$ width multipliers) on the GAP8 microcontroller using the DORY framework~\cite{dory}. We configure the GAP8 multicore cluster to operate at 1.2 V supply voltage with a clock frequency of 160 MHz as in the reference paper \cite{himax}.

Prior to deployment, we apply \gls{qat} to all models for 100 epochs, starting from their float32 checkpoints, maintaining the same optimizer (RMSprop, batch size 64) and learning rate schedule (decay factor 0.95 every 25 epochs). We empirically find that a higher initial learning rate of $1 \times 10^{-2}$ is necessary to facilitate adaptation to the quantized domain. To verify this training setup, we also apply QAT to the 1.0$\times$ static model on the original Himax dataset, obtaining a mAP of 0.547, higher than the Int8 result reported in \cite{himax} of 0.500.

Table~\ref{tab:himax_qat} reports Int8 performance on Himax EE, and deployment metrics across all configurations, including mAP, computational cost, inference efficiency in \gls{mac} operations per cycle, latency, throughput in \gls{fps}, and model size in MiB. Note that our results for static models differ slightly from those in~\cite{himax} because we use the open-source DORY compiler~\cite{dory}, rather than the proprietary toolchain of GAP8.

The static 1.0$\times$-width baseline (no EE) processes frames at 1.50 FPS with an efficiency of 4.96 \gls{mac}/cycle and 0.555 mAP. Adding the EE branch introduces minimal computational overhead: the full forward pass requires only 539 M\gls{mac}s (0.9\% increase) due to the lightweight branch architecture that adds only 0.078M parameters, maintaining nearly identical throughput at 1.48 FPS.

When early-exit is triggered at layer 9, computational cost drops dramatically to 230 M\gls{mac}s, a 57\% reduction compared to the static baseline. On the Himax EE dataset, where 39.6\% of frames exit early, the average throughput therefore reaches 1.86 FPS (24\% improvement with respect to the 1.0$\times$ static model, and comparable to 0.75$\times$) while achieving 0.540 mAP, with only a 2.7\% drop compared to the static 1.0$\times$ \gls{dnn}. The \gls{ee} branch accuracy when quantized remains high (90.9\%), with a false positive rate of 7.5\%. This demonstrates that content-based early-exit enables significant on-device acceleration on resource-constrained microcontrollers with minimal accuracy degradation for autonomous nano-drone deployment.

Finally, we also simulated the execution of our \method on the in-field data collected in~\cite{himax}. Namely, we asked the authors to provide us with the raw video from the drone camera in one of their exploration experiments\footnote{Drone flight video with onboard camera feed and detection overlay available at this link: \href{youtube.com/watch?v=BTin8g0nyko}{www.youtube.com/watch?v=BTin8g0nyko}.} and manually annotated all frames with bounding boxes for bottles and tin cans. We filtered out all boxes whose area is smaller than the smallest object in the Himax training set, as we cannot expect the DNN to predict out of distribution; hence, those frames are effectively considered empty from the model's perspective. Lastly, we ran our \gls{ee} model on the resulting data. The results show that 71.7\% of all frames are effectively empty. Our \gls{ee} model identifies them with a precision of 85.48\%. As a result, the \gls{ee} branch exits early on 54.83\% of all frames (1288 out of 2349), yielding an average throughput of 2.07 FPS, a 37.7\% improvement over a standard static MobileNetV2-SSD (1.50 FPS).

\subsection{Cityscapes results}

As a last experiment, to evaluate the generalizability of \method to larger and more complex datasets, we applied it to the patched version of Cityscapes described in Sec.~\ref{sec:setup}. The \gls{hpo} identifies layer 10 as the optimal \gls{ee} branch positioning, showing that the optimal depth range where \gls{ee} should be inserted remains similar across datasets.

\begin{table}[t]
\centering
\caption{Cityscapes results (float32).}
\label{tab:cityscapes}
\small
\setlength{\tabcolsep}{4pt}
\resizebox{\columnwidth}{!}{%
\begin{tabular}{lccccc}
\toprule
Model & mAP & Acc & Skip & \gls{mac}$_{\text{avg}}$ & \gls{mac}$_{\text{EE}}$ \\
\midrule
Static 1.0$\times$              & 0.229 & -    & -    & 1807 & -   \\
\gls{ee}, Stage 4 ($\ell=10$)  & 0.204 & 82.3 & 26.7 & 1588 & 855 \\
\bottomrule
\multicolumn{6}{p{0.95\columnwidth}}{\footnotesize Acc: EE branch accuracy, Skip: early-exited images [Unit = \%]; \gls{mac}$_{\text{avg}}$: dataset average. \gls{mac}$_{\text{EE}}$: early-exit path cost (backbone + EE branch) [Unit = M $(10^6)$]} \\
\end{tabular}
}
\end{table}

Table~\ref{tab:cityscapes} reports the results in terms of mAP and total \gls{mac} operations, comparing \method with a static 1.0$\times$ model.
Our model achieves a 12\% average computational reduction (1588 vs 1807 M\gls{mac}s) by exiting early on 26.7\% of tiles. When \gls{ee} triggers, inference costs only 855 M\gls{mac}s, a 53\% reduction compared to the full pass (1855 M\gls{mac}s), which itself incurs a modest 48 M\gls{mac}s overhead over the static baseline due to the additional EE branch. The higher absolute costs compared to Himax EE reflect the larger input resolution ($512 \times 512$ vs $240 \times 320$), resulting in approximately $3.5\times$ more M\gls{mac}s overall.

Unlike on Himax EE, the mAP of the \gls{ee} model (0.204) is slightly lower than the static baseline (0.229). This difference likely reflects the increased complexity of Cityscapes: with 8 object categories across diverse urban scenes and a higher object density per image, the empty-frame detection task is more challenging, as evidenced by the lower \gls{ee} branch accuracy. 
The lower skip rate (26.7\% vs 39.8\% on Himax \gls{ee}) further limits computational savings, although the latter could increase substantially in real-world deployments, where empty frames are more frequent.
Nonetheless, these results demonstrate that \method successfully generalizes to larger, multi-class detection scenarios, offering viable new trade-offs between accuracy and efficiency in more demanding settings.

\section{Conclusions}
\label{sec:conclusions}

We have presented \method, a simple yet effective method to improve the average \gls{od} throughput in constrained embedded devices such as nano-drones, with minimal accuracy drops. With experiments on two different datasets, we have shown that our solution provides multiple Pareto-optimal, runtime-switchable operating points in terms of mAP vs complexity, while incurring minimal memory overheads. Our future works will encompass ways to improve the \gls{ee} accuracy, e.g., by tuning the threshold $\tau$ dynamically based on previously processed frames.

\section*{Acknowledgments}
We thank the authors of \cite{himax} for providing us with pre-trained model checkpoints for comparisons, and raw drone videos.

This publication is part of the project PNRR-NGEU which has received
funding from the MUR – DM 118/2023.

{
    \small
    \bibliographystyle{ieeenat_fullname}
    \bibliography{references}
}

\end{document}